\documentclass[a4paper]{article}
\usepackage[a4paper,margin=4cm]{geometry}
\usepackage{graphicx}
\usepackage{amsmath}
\usepackage{amssymb}
\usepackage{xspace}
\usepackage{xcolor}
\usepackage{booktabs,arydshln}
\usepackage{makecell}
\usepackage{hyperref}
\usepackage{algorithm2e}

\hypersetup{colorlinks = true , urlcolor=blue} 

\makeatletter
\def\adl@drawiv#1#2#3{%
        \hskip.5\tabcolsep
        \xleaders#3{#2.5\@tempdimb #1{1}#2.5\@tempdimb}%
                #2\z@ plus1fil minus1fil\relax
        \hskip.5\tabcolsep}
\newcommand{\cdashlinelr}[1]{%
  \noalign{\vskip\aboverulesep
           \global\let\@dashdrawstore\adl@draw
           \global\let\adl@draw\adl@drawiv}
  \cdashline{#1}
  \noalign{\global\let\adl@draw\@dashdrawstore
           \vskip\belowrulesep}}
\makeatother

\newlength{\Oldarrayrulewidth}
\newcommand{\Cline}[2]{%
  \noalign{\global\setlength{\Oldarrayrulewidth}{\arrayrulewidth}}%
  \noalign{\global\setlength{\arrayrulewidth}{#1}}\cline{#2}%
  \noalign{\global\setlength{\arrayrulewidth}{\Oldarrayrulewidth}}}

\usepackage{multirow}

\usepackage{xr}

\makeatletter

\usepackage{glossaries}
\glsdisablehyper

\newacronym{ai}{AI}{Artificial Intelligence}
\newacronym{dl}{DL}{Deep Learning}
\newacronym{dnn}{DNN}{Deep Neural Network}
\newacronym{lrp}{LRP}{Layer-wise Relevance Propagation}
\newacronym{xai}{XAI}{eXplainable Artificial Intelligence}
\newacronym{crp}{CRP}{Concept Relevance Propagation}
\newacronym{amax}{ActMax}{Activation Maximization}
\newacronym{rmax}{RelMax}{Relevance Maximization}
\newacronym{auc}{AUC}{Area Under Curve}
\newacronym{aoc}{AOC}{Area Over Curve}
\newacronym{roi}{ROI}{Region of Interest}
\newacronym{sem}{SEM}{Standard Error of Mean}
\newacronym{lcrp}{L-CRP}{CRP for Localization Models}
\newacronym{rrr}{RRR}{Right for the Right Reason}
\newacronym{cdep}{CDEP}{Contextual Decomposition Explanation Penalization}
\newacronym{clarc}{ClArC}{Class Artifact Compensation}
\newacronym{aclarc}{a-ClArC}{Augmentive \gls{clarc}}
\newacronym{pclarc}{p-ClArC}{Projective \gls{clarc}}
\newacronym{ml}{ML}{Machine Learning}
\newacronym{cse}{CSE}{complete skin examination}
\newacronym{cav}{CAV}{Concept Activation Vector}
\newacronym{spray}{SpRAy}{Spectral Relevance Analysis}
\newacronym{iterrev}{IterRev}{Iteratively Revealing and Revising Spurious Model Behavior}
\newacronym{r2r}{R2R}{Reveal to Revise}
\newacronym{xil}{XIL}{eXplanatory Interactive Learning}

\newcommand{\x}{{\mathbf{x}}\xspace}

\DeclareRobustCommand\onedot{\futurelet\@let@token\@onedot}
\def\@onedot{\ifx\@let@token.\else.\null\fi\xspace}

\def\ie{\emph{i.e}\onedot}

\def\wrt{w.r.t\onedot}

\title{
\textbf{Reveal to Revise: \\An Explainable AI Life Cycle for Iterative Bias Correction of Deep Models}}
\author{Frederik Pahde$^{1,\ast}$ \and
Maximilian Dreyer$^{1,\ast}$ \and
Wojciech Samek$^{1,2,3,\dagger}$ \and
Sebastian Lapuschkin$^{1,\dagger}$}
\date{\footnotesize $^1$Fraunhofer Heinrich-Hertz-Institute, 10587 Berlin, Germany \\
$^2$Technische Universität Berlin, 10587 Berlin, Germany \\
$^3$BIFOLD – Berlin Institute for the Foundations of Learning and Data, 10587 Berlin, Germany\\
$^\dagger$ corresponding authors: \texttt{\{wojciech.samek,sebastian.lapuschkin\}@hhi.fraunhofer.de}\\
$^\ast$ contributed equally}

\begin{document}
\maketitle

\begin{abstract} State-of-the-art machine learning models often learn spurious correlations embedded in the training data. 
This poses risks when deploying these models for high-stake decision-making, such as in medical applications like skin cancer detection.
To tackle this problem, we propose \gls{r2r}, 
a framework entailing the entire \gls{xai} life cycle, 
enabling practitioners to iteratively identify, mitigate, and (re-)evaluate spurious model behavior with a 
minimal amount of human interaction.
In the first step (1),
\gls{r2r} \emph{reveals} model weaknesses by finding outliers in attributions or through inspection of latent concepts learned by the model.
Secondly (2),
the responsible artifacts are \emph{detected} and spatially \emph{localized} in the input data,
which is then leveraged to (3) \emph{revise} the model behavior.
Concretely, we apply the methods of RRR, CDEP and ClArC for model correction,
and (4) \mbox{(re-)evaluate} the model's performance and remaining sensitivity towards the artifact.
Using two medical benchmark datasets for Melanoma detection and bone age estimation,
we apply our \gls{r2r} framework to VGG, ResNet and EfficientNet architectures
and thereby reveal and correct real dataset-intrinsic artifacts, as well as synthetic variants in a controlled setting. 
Completing the XAI life cycle,
we demonstrate multiple \gls{r2r} iterations to mitigate different biases. 
Code is available on \url{https://github.com/maxdreyer/Reveal2Revise}.
%
\end{abstract}

\section{Introduction}

    \glspl{dnn} have successfully been applied in research and industry for a multitude of complex tasks. 
    This includes various medical applications for which \glspl{dnn} have even shown to be superior to medical experts, such as with Melanoma detection \cite{brinker2019deep}.
    However, 
    the reasoning of these highly complex and non-linear models is generally not transparent \cite{rudin2019stop,samek2021explaining}, 
    and as such, 
    their decisions may be biased towards unintended or undesired features \cite{stock2018convnets,lapuschkin2019unmasking,anders2022finding}. 
    Particularly in high-stake decision processes, such as medical applications, unreliable or poorly understood model behavior may pose severe security risks.

    The field of \gls{xai} brings light into the black boxes of \glspl{dnn} and provides a better understanding of their decision processes.
    As such, local \gls{xai} methods reveal (input) features that are most relevant to
    a
    model, 
    which, for image data, can be presented as heatmaps. 
    In contrast,
    global \gls{xai} methods (e.g, \cite{kim2018interpretability,lapuschkin2019unmasking}) reveal general prediction strategies employed or features encoded by a model, 
    which is necessary for the identification and understanding of systematic (mis-)behavior.
    Acting on the insights from explanations, 
    various methods have been introduced to correct for undesired model behavior \cite{weber2022beyond}.
    While multiple approaches exist for either \emph{revealing} or \emph{revising} model biases,
    only few combine both steps, to be applicable as a framework.
    Such frameworks,
    however, either rely heavily on human feedback~\cite{teso2019explanatory,schramowski2020making},
    are limited to specific bias types \cite{anders2022finding},
    or require labor-intensive annotations for both model evaluation and correction~\cite{schramowski2020making,kim2019learning}.

    \begin{figure}[t]
            \includegraphics[width=\textwidth]{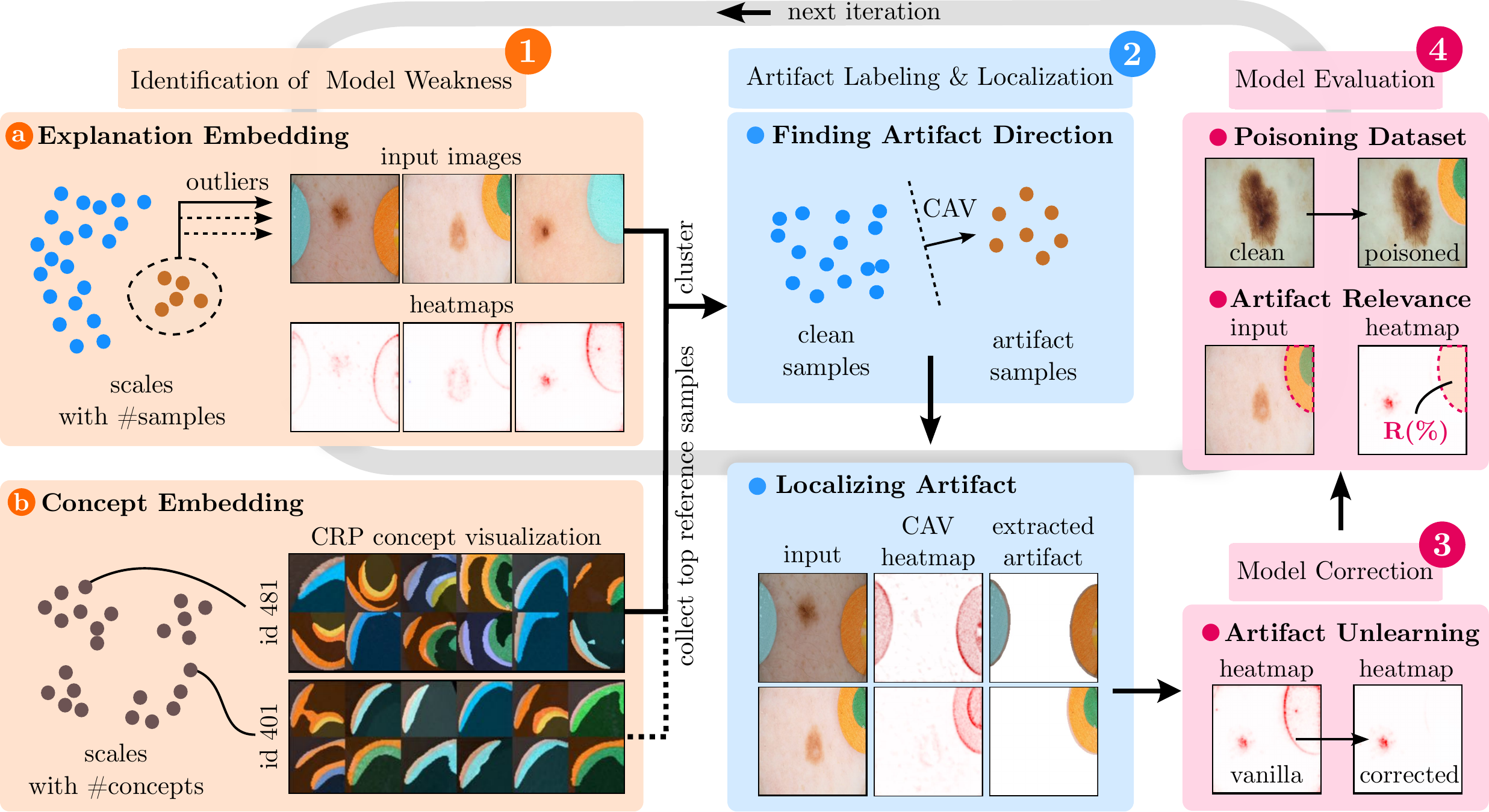}
            \caption{Our R2R life cycle for \emph{revealing} and \emph{revising} spurious behavior of any pre-trained DNN.
            Firstly,
            we identify model weaknesses by finding either outliers in explanations using SpRAy (1a) or suspicious concepts using CRP concept visualizations (1b).
            Secondly (2),
            SpRAy clusters or collecting the top reference samples allows us to label artifactual samples and to compute an artifact CAV,
            which we use to model and localize the artifact in latent and input space, respectively.
            At this point,
            the artifact localization can be leveraged for (3) model correction,
            and (4) to evaluate the model's performance on a poisoned test set and measure its remaining attention on the artifact.
            }
            \label{fig:xai_lifecycle}
    \end{figure}
    
    To that end, we propose \glsdesc{r2r} (\gls{r2r}), 
    an iterative \gls{xai} life cycle requiring low amounts of human interaction that consists of four phases, illustrated in Fig.~\ref{fig:xai_lifecycle}.
    Specifically, 
    \gls{r2r} allows to first (1) identify spurious model behavior
    and secondly, to (2) label and localize artifacts in an automated fashion.
    The generated annotations are then leveraged to (3) correct and (4) (re-)evaluate the model,
    followed by a repetition of the entire life cycle if required.
    For \emph{revealing} model bias,
    we propose two orthogonal \gls{xai} approaches:
    While \gls{spray}~\cite{lapuschkin2019unmasking} automatically finds outliers in model explanations (potentially caused by the use of spurious features),
    \gls{crp}\cite{achtibat2022towards} precisely communicates the globally learned concepts of a \gls{dnn}.
    For model \emph{revision},
    we apply and compare the methods of \gls{clarc}~\cite{anders2022finding}, \gls{cdep}~\cite{rieger2020interpretations} and \gls{rrr}~\cite{ross2017right},
    penalizing attention on artifacts via ground truth masks automatically generated in step (2).
    The artifact masks are further used for evaluation on a poisoned test set and to measure the remaining attention on the bias.
    We demonstrate the applicability and high automation of \gls{r2r} on two medical tasks, 
    including Melanoma detection and bone age estimation, using the VGG-16, ResNet-18 and EfficientNet-B0 \gls{dnn} architectures. 
    In our experiments, we correct model behavior \wrt dataset-intrinsic, as well as synthetic artifacts in a controlled setting.
    Lastly, 
    we showcase the \gls{r2r} life cycle through multiple iterations, unveiling and unlearning different biases.

\section{Related Work}

    The majority of related works introduce methods to either identify spurious behavior \cite{lapuschkin2019unmasking,achtibat2022towards},
    or to align the model behavior with pre-defined priors \cite{ross2017right,rieger2020interpretations},
    with only few combining both, such as the \gls{xil} framework \cite{teso2019explanatory} or the approach introduced by Anders et al.~\cite{anders2022finding}.
    The former is based on presenting individual local explanations to a human,
    who, if necessary, provides feedback used for model correction \cite{teso2019explanatory,schramowski2020making}.
    However,
    studying individual predictions is slow and labor-extensive, limiting its practicability.
    In contrast,
    the authors of \cite{anders2022finding} use \gls{spray} \cite{lapuschkin2019unmasking} for the detection of spurious model behavior and labeling of artifactual samples.
    In addition to \gls{spray}, 
    we suggest to study latent features of the model via \gls{crp} concept visualizations \cite{achtibat2022towards} as a tool for more fine-grained model inspection,
    catching systematic misbehavior which would not be visible through \gls{spray} clusters.

    Most model correction methods require dense annotations, such as labels for artifactual samples or artifact localization masks,
    which are either crafted heuristically or by hand \cite{rieger2020interpretations,kim2019learning}.
    In our \gls{r2r} framework, 
    we automate the annotation by
    following \cite{anders2022finding} for data labeling through \gls{spray} outlier clusters,
    or by collecting the most representative samples of bias concepts according to \gls{crp}.
    The spatial artifact localization is further automated by computing artifact heatmaps as outlined in Section~\ref{sec:art_localization},
    thereby considerably easing the step from bias identification to correction.

    Existing works for model correction measure the performance on the original or clean test set,
    with corrected models often showing an improved generalization \cite{kim2019learning,rieger2020interpretations}.
    A more targeted approach for measuring the artifact's influence is the evaluation on poisoned data \cite{schramowski2020making},
    for which \gls{r2r} is well suited by using its localization scheme to first extract artifacts and to then poison clean test samples.
    By precisely localizing artifacts, \gls{r2r} further allows to measure the model's attention on an artifact through attribution heatmaps.


\section{Reveal to Revise Framework}

    Our \emph{\glsdesc{r2r}} (\gls{r2r}) framework comprises the entire \gls{xai} life cycle, 
    including methods for (1) the identification of model bias,
    (2) artifact labeling and localization,
    (3) the correction of detected misbehavior, and (4) the evaluation of the improved model.
    To that end, we now describe the methods used for \gls{r2r}.

    \subsection{Data Artifact Identification and Localization}
        \label{sec:art_localization}
        
        The identification of spurious data artifacts using \gls{crp} concept visualizations or \gls{spray} clusters is firstly described,
        followed by our artifact localization approach.
        
        \subsubsection{CRP Concept Visualizations}

            \gls{crp}~\cite{achtibat2022towards} combines global concept visualization techniques with local feature attribution methods.
            This provides an understanding of the relevance of latent concepts for a prediction and their localization in the input.
            In this work, we use \gls{lrp}~\cite{bach2015pixel} for feature attribution under \gls{crp} and for heatmaps in general, however, other local \gls{xai} methods can be used as well.
            Jointly with Relevance~Maximization~\cite{achtibat2022towards}, \gls{crp} is well suited for the identification of spurious concepts
            by precisely narrowing down the input parts that have been most relevant for model inference,
            as shown in Fig.~\ref{fig:xai_lifecycle} (\emph{bottom left}) for band-aid concepts,
            where irrelevant background is overlaid with black semi-transparent color.
            The collection of top-ranked reference samples for spurious concepts allows us to label artifactual data.

            
            
        
        \subsubsection{Explanation Outliers Through SpRAy}

            Alternatively, 
            \gls{spray} \cite{lapuschkin2019unmasking} is a strategy to find outliers in local explanations,
            which are likely to stem from spurious model behavior, such as the use of a Clever Hans features, \ie, features correlating with a certain class that are unrelated to the actual task.
            Following \cite{lapuschkin2019unmasking,anders2022finding},
            we apply \gls{spray} by clustering latent attributions computed through \gls{lrp}.
            The \gls{spray} clusters then naturally allow us to label data containing the bias.

        \subsubsection{Artifact Localization}

           We automate artifact localization by using a \gls{cav} $\mathbf{h}_l$ to model the artifact in latent space of a layer $l$, representing the direction from artifactual to non-artifactual samples obtained from a linear classifier.
            %
            The artifact localization is given by a modified backward pass with \gls{lrp} for an artifact sample $\x$,
            where we initialize the relevances $\mathbf{R}_l(\x)$ at layer $l$ as  
            \begin{equation}\label{eq:localization}
                \mathbf{R}_l(\x) = \mathbf{a}_l(\x) \circ \mathbf{h}_l\ 
            \end{equation}
            with activations $\mathbf{a}_l$ and element-wise multiplication operator $\circ$.
            This is equivalent to explaining the output from the linear classifier given as $\mathbf{a}_l(\x) \cdot \mathbf{h}_l$.
            The resulting \gls{cav} heatmap can be further processed into a binary mask to crop out the artifact from any corrupted sample, as illustrated in Fig.~\ref{fig:xai_lifecycle} (\emph{bottom center}).
    
    \subsection{Methods for Model Correction}

            In the following, we present the methods used for mitigating model biases.
    
        \subsubsection{ClArC for Latent Space Correction} 
            Methods from the \gls{clarc} framework correct model (mis-)behavior \wrt an artifact by modeling its direction $\mathbf{h}$ in latent space using \glspl{cav}~\cite{kim2018interpretability}. 
            The framework consists of two methods, namely \gls{aclarc} and \gls{pclarc}. 
            While \gls{aclarc} adds $\mathbf{h}_l$ to the activations $\mathbf{a}_l$ of layer $l$ for all samples in a fine-tuning phase, 
            hence teaching the model to be invariant towards that direction, 
            \gls{pclarc} suppresses the artifact direction during the test phase and does not require any fine-tuning.
            More precisely,
            the perturbed activations $\mathbf{a}_l'$ are given by
            \begin{equation}
                \mathbf{a}_l' (\x) = \mathbf{a}_l (\x) + \gamma(\x) \mathbf{h}_l
            \end{equation}
            with perturbation strength $\gamma(\x)$ dependent on input $\x$.
            Parameter $\gamma(\x)$ is chosen such that the activation in direction of the \gls{cav} is as high as the average value over non-artifactual or artifactual samples for \gls{pclarc} or \gls{aclarc}, respectively.
            
        \subsubsection{\gls{rrr} and \gls{cdep} for Correction through Prior Knowledge}
            Model correction using \gls{rrr} \cite{ross2017right} or \gls{cdep} \cite{rieger2020interpretations} is based on an additional $\lambda$-weighted loss term (besides the cross-entropy loss $\mathcal{L}_\text{CE}$) for neural network training
            that aligns the use of features by the model $f_\theta$, described by an explanation $\text{exp}_\theta$, to a defined prior explanation $\text{exp}_{\text{prior}}$.
            %
            The authors of \gls{rrr} propose to penalize the model's attention on unfavorable artifacts using the input gradient \wrt the cross-entropy loss, 
            leading to
            \begin{equation}
                \mathcal{L}_\text{RRR}\left( \text{exp}_\theta(\x), \text{exp}_{\text{prior}}(\x)\right) 
                = {\| \boldsymbol{\nabla}_\x \mathcal{L}_\text{CE} \left(f_\theta(\x), y_\text{true}\right) \circ \mathbf{M}_\text{prior}(\x)   \|_2}^2
            \end{equation}
            with a binary mask $\mathbf{M}_\text{prior}(\x)$ localizing an artifact and class label $y_\text{true}$.
            We further adapt the \gls{rrr} loss to increase stability in regard to the high variance of \gls{dnn} gradients by using the cosine similarity (instead of L2-norm) and compute the gradient \wrt the predicted logit $p$, leading to
            \begin{equation}
                \mathcal{L}_\text{RRR}\left( \text{exp}_\theta(\x), \text{exp}_{\text{prior}}(\x)\right) 
                = \frac{ |\boldsymbol{\nabla}_\x f_\theta(\x)_p| \cdot \mathbf{M}_\text{prior}(\x)   }
                {\|\boldsymbol{\nabla}_\x f_\theta(\x)_p\|_2\|\mathbf{M}_\text{prior}(\x)\|_2}.
            \end{equation}
        
            Alternatively,
            \gls{cdep}~\cite{rieger2020interpretations} proposes to use CD \cite{murdoch2018beyond} importance scores $\boldsymbol{\beta}(\x_s)$ for a feature subset $\x_s$ based on the forward pass instead of gradient to align the model's attention.
            Penalizing artifact features via masked input $\x_M$ results in 
            \begin{equation}
                \mathcal{L}_\text{CDEP}\left( \text{exp}_\theta(\x), \text{exp}_{\text{prior}}(\x)\right) 
                = \left\| \frac{e^{\boldsymbol{\beta}(\x_M)}}{e^{\boldsymbol{\beta}(\x_M)} + e^{\boldsymbol{\beta}(\x - \x_M)}} \right\|_1\,.
            \end{equation}

\section{Experiments} \label{sec:exp}

    The experimental section is divided into the two parts of (1) identification, mitigation and evaluation of spurious model behavior with various correction methods
    and (2) showcasing the whole \gls{r2r} framework in an iterative fashion.

    \subsection{Experimental Setup} \label{sec:exp:details}
    
        We train VGG-16~\cite{simonyan2014very}, ResNet-18~\cite{he2016deep} and EfficientNet-B0~\cite{tan2019efficientnet} models on the ISIC 2019 dataset \cite{tschandl2018ham10000,codella2018skin,combalia2019bcn20000} for skin lesion classification
        and Pediatric Bone Age dataset~\cite{halabi2019rsna} for bone age estimation based on hand radiographs. 
        Besides evaluating our methodology on data-intrinsic artifacts occurring in these datasets,
        we artificially insert an artifact into data samples in a controlled setting. 
        Specifically, we insert a ``Clever Hans'' text (shown in Fig.~\ref{fig:exp:artifacts_overview}) into 
        a subset of training samples of one specific class.
        See Appendix~\ref{app:exp:details} for additional experiment details.
    
        \begin{figure}[t]
            \includegraphics[width=\textwidth]{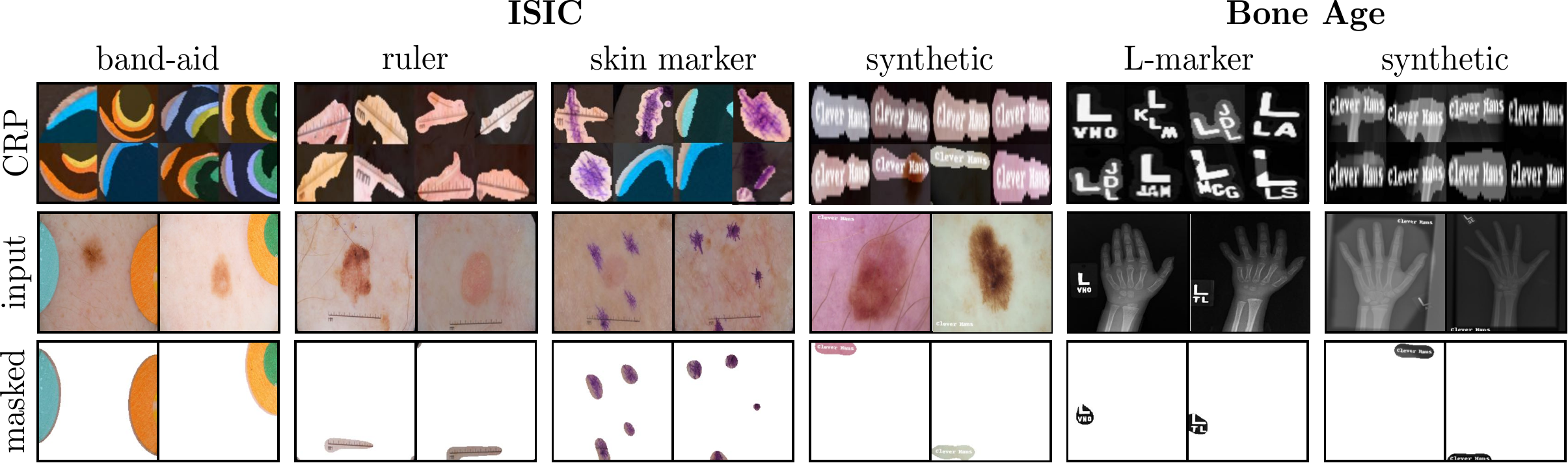}
            \caption{Overview of artifacts with \gls{crp} visualization of corresponding concepts (\emph{top}), input samples (\emph{middle}), and cropped out artifacts (\emph{bottom}) using our artifact localization method. Shown are band-aid, ruler, skin marker, and synthetic artifacts for the ISIC dataset, as well as ``L''-marker and synthetic artifacts for the Bone Age dataset. } \label{fig:exp:artifacts_overview}
        \end{figure}
        
    \subsection{Revealing and Revising Spurious Model Behavior} \label{sec:exp:revealing}

        \paragraph{Revealing Bias:} 
        In the first step of the \gls{r2r} life cycle, 
        we can reveal the use of several artifacts by the examined models,
        including the well-known band-aid, 
        ruler and skin marker \cite{cassidy2022analysis} and our synthetic Clever Hans for the ISIC dataset,
        as shown in Fig.~\ref{fig:exp:artifacts_overview} for VGG-16.
        Here,
        we show concept visualizations and cropped out artifacts based on our automatic artifact localization scheme described in Section~\ref{sec:art_localization}.
        The ``band-aid'' use can be further identified via \gls{spray}, as illustrated in Fig.~\ref{fig:iterative_correction_qualitative} (\emph{right}).
        Examplary artifact \gls{cav} heatmaps for all data-intrinsic artifacts are given in Appendix~\ref{app:sec:additional_results:localization}.

        Besides the synthetic Clever Hans for bone age classification,
        we encountered the use of ``L'' markings,
        resulting from physical lead markers placed by radiologist to specify the anatomical side.
        Interestingly,
        the ``L'' markings are larger for hands of younger children, 
        as all hands are scaled to similar size \cite{halabi2019rsna}, 
        offering the model to learn a shortcut by estimating the bone age based on the relative size of the ``L'' markings, instead of valid features.
        While we revealed the ``L'' marking bias using \gls{crp},
        we did not find corresponding \gls{spray} clusters,
        underlining the importance of both approaches for model investigation.
        

        
    

\paragraph{Revising Model Behavior:} Having revealed spurious behavior,
we now revise the models, beginning with model correction.
Specifically,
we correct for the band-aid, ``L'' markings as well as synthetic artifacts.
The skin marker and ruler artifacts are corrected for in iterative fashion in Section~\ref{sec:exp:iterative}.
For all methods (\gls{rrr}, \gls{cdep}\footnote{CDEP is not applied to EfficientNets, as existing implementations are incompatible.}  
and \gls{clarc}), including a \emph{Vanilla} model without correction, we fine-tune the models' last dense layers for 10 epochs.
Note that both \gls{rrr} and \gls{cdep} require artifact masks to unlearn the undesired behavior.
As part of \gls{r2r}, we propose measures to automate this step by using the artifact localization strategy described in Section~\ref{sec:art_localization}. 
Further note,
that once generated,
artifact localizations can be used for \emph{all} investigated models.
See Appendix~\ref{app:exp:details} for additional fine-tuning details.

    \begin{table}[t]\centering
        \caption{Model correction results for two ISIC dataset artifacts (band-aid$\,|\,$synthetic). Arrows indicate whether low ($\downarrow$) or high ($\uparrow$) scores are better with best scores bold.}\label{tab:model_correction_isic}
\begin{tabular}{@{}c@{\hspace{1em}}lc@{\hspace{0.5em}}c@{\hspace{0.5em}}c@{\hspace{1em}}c@{\hspace{0.5em}}c@{}}
\toprule

&         
& $\downarrow$ artifact 
&  \multicolumn{2}{c}{$\uparrow$ F1 (\%)}
&  \multicolumn{2}{c}{$\uparrow$ accuracy (\%)}  \\ 
architecture & method  & relevance (\%) & \textit{poisoned} & \textit{original} & \textit{poisoned} & \textit{original} \\ 
\midrule

\multirow{5}{*}{VGG-16} 
    &\emph{Vanilla}&                    ${45.5}\,|\,{76.3}$ &         ${59.7}\,|\,{7.7}\;\;$ &  ${73.9}\,|\,{79.0}$ &              ${71.5}\,|\,{19.1}$ &        ${80.1}\,|\,{86.9}$ \\[.15ex] \Cline{.1pt}{2-7}\rule{0pt}{2.5ex}    
    
    &           RRR &                    $\mathbf{14.3}\,|\,\mathbf{12.0}$ &        $\mathbf{64.2}\,|\,\mathbf{39.2}$ &  $\mathbf{}71.8\,|\,\mathbf{}77.7$ &              $\mathbf{74.4}\,|\,\mathbf{32.4}$ &        $\mathbf{}78.0\,|\,\mathbf{}85.4$ \\
    &          CDEP &                    $\mathbf{}23.7\,|\,\mathbf{}78.4$ &         $\mathbf{}62.8\,|\,\mathbf{}7.2\;\;$ &  $\mathbf{}\:73.9\,|\,\mathbf{79.0}$ &              $\mathbf{}72.3\,|\,\mathbf{}18.9$ &        $\mathbf{}80.2\,|\,\mathbf{}86.9$ \\
    &       \gls{pclarc} &                    $\mathbf{}41.9\,|\,\mathbf{}76.1$ &         $\mathbf{}61.8\,|\,\mathbf{}7.6\;\;$ &  $\mathbf{74.0}\,|\,\mathbf{}78.1\:$ &              $\mathbf{}73.0\,|\,\mathbf{}19.1$ &        $\mathbf{80.3}\,|\,\mathbf{}85.4\:$ \\
    &       \gls{aclarc} &                    $\mathbf{}42.8\,|\,\mathbf{}75.5$ &        $\mathbf{}62.4\,|\,\mathbf{}12.5$ &  $\mathbf{}70.3\,|\,\mathbf{}76.5$ &              $\mathbf{}73.1\,|\,\mathbf{}21.0$ &        $\:\mathbf{}78.4\,|\,\mathbf{88.9}$ \\
    \midrule
    
\multirow{5}{*}{ResNet-18} 
    &\emph{Vanilla} &                    ${33.1}\,|\,{37.6}$ &        ${68.2}\,|\,{39.0}$ &  ${79.1}\,|\,{82.1}$ &              ${76.8}\,|\,{35.6}$ &        ${83.3}\,|\,{89.5}$ \\ \Cline{.1pt}{2-7}\rule{0pt}{2.5ex} 
    &           RRR &                    $\:\mathbf{}30.3\,|\,\mathbf{16.9}$ &        $\:\mathbf{}70.4\,|\,\mathbf{70.4}$ &  $\mathbf{79.7}\,|\,\mathbf{}79.1\:$ &              $\:\mathbf{}77.1\,|\,\mathbf{75.7}$ &        $\mathbf{83.4}\,|\,\mathbf{}84.8\:$ \\
    &          CDEP &                    $\mathbf{25.4}\,|\,\mathbf{}22.2\:$ &        $\mathbf{71.5}\,|\,\mathbf{}60.9\:$ &  $\mathbf{}75.9\,|\,\mathbf{}81.6$ &              $\mathbf{77.5}\,|\,\mathbf{}59.4\:$ &        $\mathbf{}81.5\,|\,\mathbf{}87.9$ \\
    &       \gls{pclarc} &                    $\mathbf{}32.0\,|\,\mathbf{}33.6$ &        $\mathbf{}69.2\,|\,\mathbf{}38.9$ &  $\:\mathbf{}78.3\,|\,\mathbf{81.8}$ &              $\mathbf{}75.9\,|\,\mathbf{}34.4$ &        $\:\mathbf{}82.5\,|\,\mathbf{89.1}$ \\
    &       \gls{aclarc} &                    $\mathbf{}32.9\,|\,\mathbf{}38.4$ &        $\mathbf{}70.1\,|\,\mathbf{}52.9$ &  $\mathbf{}78.3\,|\,\mathbf{}80.5$ &              $\mathbf{}76.2\,|\,\mathbf{}45.3$ &        $\mathbf{}81.1\,|\,\mathbf{}88.9$ \\
    \midrule
\multirow{4}{*}{\shortstack[c]{Efficient-\\Net-B0}} 
    &\emph{Vanilla} &                    ${45.6}\,|\,{63.9}$ &        ${72.2}\,|\,{38.8}$ &  ${81.8}\,|\,{84.7}$ &              ${80.1}\,|\,{30.2}$ &        ${85.4}\,|\,{90.8}$ \\ [0.15ex] \Cline{.1pt}{2-7}\rule{0pt}{2.5ex} 
    &           RRR &                    $\mathbf{34.5}\,|\,\mathbf{24.6}$ &        $\mathbf{74.0}\,|\,\mathbf{}65.8\:$ &  $\mathbf{}81.3\,|\,\mathbf{}83.3$ &              $\mathbf{}80.1\,|\,\mathbf{}65.9$ &        $\mathbf{}84.6\,|\,\mathbf{}89.8$ \\
    &       \gls{pclarc} &                    $\mathbf{}41.3\,|\,\mathbf{}62.5$ &        $\mathbf{}73.1\,|\,\mathbf{}38.7$ &  $\mathbf{82.0}\,|\,\mathbf{84.4}$ &              $\mathbf{80.4}\,|\,\mathbf{}29.8\:$ &        $\mathbf{85.5}\,|\,\mathbf{90.5}$ \\
    &       \gls{aclarc} &                    $\mathbf{}45.7\,|\,\mathbf{}65.6$ &        $\:\mathbf{}72.7\,|\,\mathbf{72.4}$ &  $\mathbf{}81.8\,|\,\mathbf{}81.4$ &              $\:\mathbf{}80.1\,|\,\mathbf{79.4}$ &        $\mathbf{}84.9\,|\,\mathbf{}87.3$ \\
    \bottomrule
\end{tabular}
\end{table}
We evaluate the effectiveness of model corrections based on two metrics: the attributed fraction of relevance to artifacts
and prediction performance on both the original and a poisoned test set (in terms of F1-score and accuracy).
Whereas in the synthetic case, we simply insert the artifact into all samples to poison the test set,
data-intrinsic artifacts are cropped from random artifactual samples using our artifact localization strategy.
Note
that artifacts might overlap clinically informative features in poisoned samples, 
limiting the comparability of \emph{poisoned} and \emph{original} test performance.
As shown in Tab.~\ref{tab:model_correction_isic} (ISIC 2019) and 
Appendix~\ref{app:sec:additional_results} (Bone Age),
we are generally able to improve model behavior with all methods.
The only exception is the synthetic artifact for VGG-16, where only \gls{rrr} mitigates the bias to a certain extent,
indicating that the artifact signal is too strong for the model.
Here, fine-tuning only the last layer is not sufficient to learn alternative prediction strategies.

Interestingly, 
despite successfully decreasing the models' output sensitivity towards artifacts, applying  
\gls{aclarc} barely decreases the relevance attributed to artifacts in input space.
This might result from \gls{clarc} methods not directly penalizing the use of artifacts,
but instead encouraging the model to develop alternative prediction strategies.
Overall,
\gls{rrr} yields the most consistent results, 
constantly reducing the artifact relevance while increasing the model performance on poisoned test sets.
Both observations are underlined by heatmaps for revised models in Fig.~\ref{app:fig:qualititave_correction_results}~(Appendix~\ref{app:sec:additional_results}), 
where \gls{rrr} and \gls{cdep} visibly reduce the model attention on the artifacts.



\subsection{Iterative Model Correction with R2R} \label{sec:exp:iterative}
Showcasing the full \gls{r2r} life cycle (as shown in Fig.~\ref{fig:xai_lifecycle}),
we now perform multiple \gls{r2r} iterations, revealing and revising undesired model behavior step by step.
Specifically, we successively correct the VGG-16 model \wrt the skin marker, band-aid, and ruler artifacts discovered in Section~\ref{sec:exp:revealing} using \gls{rrr}. 
In order to prevent the model from re-learning previously unlearned artifacts,
we keep the previous artifact-specific \gls{rrr} losses intact.
\begin{figure}[t] \centering
        \includegraphics[width=0.91\textwidth]{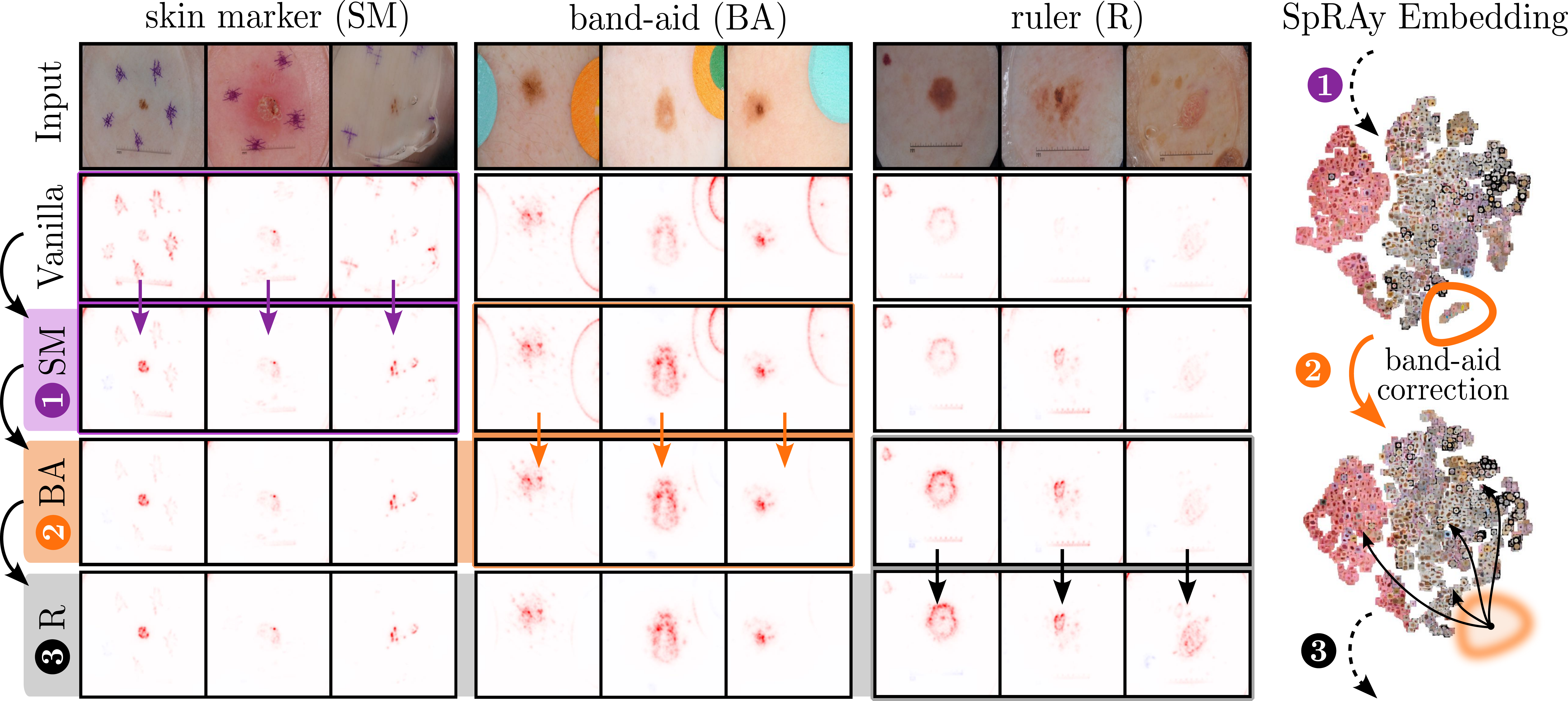}
        \caption{The effect of iterative model correction on relevances attributed to artifacts for each iteration (\emph{left}) and the band-aid artifact cluster from \gls{spray},
        which dissipates after its correction step (\emph{right}). 
        }
        \label{fig:iterative_correction_qualitative}
    \end{figure}
Thus,
we are able to correct for all artifacts, 
with evaluation results given in Tab.~\ref{tab:iterative_model_correction}, applying the same metrics as in Section~\ref{sec:exp:revealing}.
In Fig.~\ref{fig:iterative_correction_qualitative}, 
we show exemplary attribution heatmaps for all artifacts after each iteration.
While there are large amounts of relevance on all artifacts initially, 
it can successfully be reduced in the according iterations to correct the model behavior \wrt skin marker (SM), band-aids (BA), and rulers (R).
It is to note,
that correcting for the skin marker also (slightly) improved the model \wrt other artifacts,
which might result from corresponding latent features that are not independent, as shown by \gls{crp} visualizations in Fig.~\ref{fig:exp:artifacts_overview} for skin marker.
Moreover, 
we show the \gls{spray} embedding of training samples after the first iteration in Fig.~\ref{fig:iterative_correction_qualitative}~(\emph{right}), revealing an isolated cluster with samples containing the band-aid artifact, 
which dissipates after the correction step.

\begin{table}[t]\centering
\caption{Iterative R2R results for ISIC artifacts (skin marker: SM$\,|\,$band-aid: BA$\,|\,$ruler: R). Arrows show whether low ($\downarrow$)/high ($\uparrow$) scores are better, best in bold.}\label{tab:iterative_model_correction}
\resizebox{\textwidth}{!}{
\begin{tabular}{@{}c@{\hspace{0.5em}}l@{\hspace{0.5em}}c@{\hspace{0.5em}}c@{\hspace{0.5em}}c@{\hspace{0.5em}}c@{\hspace{0.5em}}c@{}}
\toprule
 R2R & corrected & $\downarrow$ artifact &  \multicolumn{2}{c}{\hspace{2.0em} $\uparrow$ F1 (\%)} &  \multicolumn{2}{c}{\hspace{1.0em} $\uparrow$ accuracy (\%)}  \\ 
iteration & artifacts  & relevance (\%) & \textit{poisoned} & \textit{original} & \textit{poisoned} & \textit{original} \\ 
\midrule 
   0 & -            & $\:\mathbf{}18.4\,|\,\:\mathbf{}45.5\,|\,\mathbf{}24.2\:$ & $\:\mathbf{}61.3\,|\,\:\mathbf{}59.7\,|\,\mathbf{}60.5\:$ & $\mathbf{}73.9$ & $\:\mathbf{}71.8\,|\,\:\mathbf{}71.5\,|\,\mathbf{}68.7\:$ & $\mathbf{80.1}$\\
   1 & SM  & $\:\mathbf{}13.1\,|\,\:\mathbf{}35.0\,|\,\mathbf{}21.3\:$ & $\:\mathbf{}61.6\,|\,\:\mathbf{}61.0\,|\,\mathbf{}60.7\:$ & $\mathbf{}73.8$ & $\:\mathbf{}72.2\,|\,\:\mathbf{}72.6\,|\,\mathbf{}68.4\:$ & $\mathbf{}80.0$\\
   2 & SM, BA     & $\mathbf{12.8}\,|\,\:\mathbf{}16.8\,|\,\mathbf{}16.8\:$ & $\:\mathbf{}61.5\,|\,\mathbf{63.6}\,|\,\mathbf{}61.1\:$ & $\mathbf{}73.9$ & $\:\mathbf{}72.3\,|\,\mathbf{74.6}\,|\,\mathbf{}68.6\:$ & $\mathbf{}79.7$\\
   3 & SM, BA, R        & $\:\mathbf{}14.6\,|\,\mathbf{15.7}\,|\,\mathbf{8.5}\;\;$ & $\mathbf{62.0}\,|\,\:\mathbf{}63.4\,|\,\mathbf{64.0}$ & $\mathbf{74.0}$ & $\mathbf{72.4}\,|\,\:\mathbf{}74.5\,|\,\mathbf{71.8}$ & $\mathbf{}79.9$\\
    \bottomrule
\end{tabular}
}
\end{table}

\section{Conclusion}
    We present \gls{r2r}, 
    an \gls{xai} life cycle to reveal and revise spurious model behavior requiring minimal human interaction via high automation.
    To \emph{reveal} model bias,
    \gls{r2r} relies on \gls{crp} and \gls{spray}. 
    Whereas \gls{spray} automatically points out Clever Hans behavior by analyzing large sets of attribution data,
    \gls{crp} allows for a fine-grained investigation of spurious concepts learned by a model.
    Moreover, \gls{crp} is ideal for large datasets, as the concept space dimension remains constant.
    By automatically localizing artifacts,
    we successfully perform model \emph{revision},
    thereby reducing attention on the artifact and leading to improved performance on corrupted data.
    When applying \gls{r2r} iteratively, 
    we did not find the emergence of new biases,
    which, however, might happen if larger parts of the model are fine-tuned or retrained to correct strong biases.
    Future research directions include the application to non-localizable artifacts,
    and addressing fairness issues in \glspl{dnn}.
    

\section*{Acknowledgements}
This work was supported by
the Federal Ministry of Education and Research (BMBF) as grants [SyReal (01IS21069B), BIFOLD (01IS18025A, 01IS18037I)];
the European Union's Horizon 2020 research and innovation programme (EU Horizon 2020) as grant [iToBoS (965221)];
the state of Berlin within the innovation support program ProFIT (IBB) as grant [BerDiBa (10174498)]; and the German Research Foundation [DFG KI-FOR 5363].


%
%

\bibliographystyle{bib}
\bibliography{mybibliography}
\newpage
\appendix

\section{Appendix}
\subsection{\emph{Reveal to Revise} Algorithm} \label{app:algorithm}
\begin{algorithm}
\DontPrintSemicolon
\caption{R2R Algorithm as outlined in Fig.~\ref{fig:xai_lifecycle}: We identify model weaknesses by finding either outliers in explanations using SpRAy (1a) or suspicious concepts using CRP concept visualizations (1b).
            Secondly (2),
            SpRAy clusters or collecting the top reference samples allows us to label artifactual samples and to compute an artifact CAV,
            which we use to model and localize the artifact in latent and input space, respectively.
            At this point,
            the artifact localization can be leveraged for (3) model correction,
            and (4) to evaluate the model's performance on a poisoned test set and measure its remaining attention on the artifact. This algorithm can be repeated until no more model weaknesses can be detected.}\label{app:alg:r2r}
\KwIn{Biased Model $f$, Training Data $X, y$}
\KwOut{Corrected Model $\hat{f}$}
$\hat{f} \gets f$\;\;
\tcc{(1) Identification of model weaknesses via \emph{SpRAy} (1a) or \emph{CRP} concpt visualizations (1b)}
$
w
\gets \text{identifyWeaknesses(}\hat{f}, X\text{)}$\;\;

\While{$w \neq \{\}$}{
    \tcc{(2) Artifact labeling and localization with CAV}
    $X_{art} \gets \text{labelArtifactualSamples(}\hat{f}, X, w\text{)}$\;
    $h_{art} \gets \text{fitCAV(}\hat{f}, X_{art}, X\text{)}$\;
   
    $M_{art} \gets \text{localizeArtifacts(}\hat{f}, h_{art}, X_{art}\text{)}$\;\;

    \tcc{(3) Model correction via RRR, CDEP or ClArC}
    \uIf{Correction Method from ClArC-Framework}{
    \tcc{Correct with artifact direction in latent space}
       $\hat{f} \gets \text{correctModel(}\hat{f}, X, h_{art}\text{)}$\;
    }
    \Else{
    \tcc{Correct with (automatically detected) artifact masks}
    $\hat{f} \gets \text{correctModel(}\hat{f}, X, M_{art}\text{)}$\;
    }\;

    \tcc{(4) (Re-)evaluate Model}
    $\text{evaluateModel(}\hat{f}, X, M_{art}, y\text{)}$\;\;

    \tcc{Repeat from (1)}
    $w \gets \text{identifyWeaknesses(}\hat{f}, X\text{)}$\;
}
\end{algorithm}

\subsection{Experimental Details} \label{app:exp:details}
    
    
    \begin{table}[h!]\centering
    \setlength{\tabcolsep}{.3em}
    \caption{Details for ISIC2019 and Pediatric Bone Age datasets. All images are resized to the same input size and normalized to mean $\mu$ and standard deviation $\sigma$. In the controlled setting, we insert a ``Clever Hans''-text with random size, position, and rotation into a chosen class with given probability $p$. Each dataset is split into training data to train the models, as well as to detect and unlearn undesired behavior,  validation data to pick optimal $\lambda$ values for correction methods, and test data for evaluation.
    }
    \resizebox{\textwidth}{!}{
    \begin{tabular}{lcccccc} \toprule
    & number  & input & norm.  &  & split size & Clever Hans\\ 
    dataset&samples &size & ($\mu/\sigma$)& classes & train/val/test &class ($p$)\\\midrule
    ISIC2019 &25331&$224\times224$& $0.5/0.5$ & \makecell{MEL, NV, BCC, AK,\\BKL, DF, VASC, SCC} &  $0.8/0.1/0.1$   &  MEL (10\%)  \\
    Bone Age &12611&$224\times224$& $0.5/0.5$& \makecell{0-46, 47-91, 92-137,\\138-182, 183-228 (months)} &  $0.8/0.1/0.1$ & 0-46 (50\%)  \\ \bottomrule
    \end{tabular}
    }
    \end{table}
    \begin{table}[h!]\centering
    \setlength{\tabcolsep}{.5em}
    \caption{Training details for examined architectures. Weights are pre-trained from the PyTorch model zoo. The learning rate is divided by 10 after 50 and 80 epochs.}
    \resizebox{\textwidth}{!}{
    \begin{tabular}{lcccc} \toprule
    {architecture}& optimizer & loss & epochs (ISIC/Bone Age) & initial learning rate  \\ \midrule
    VGG-16 & SGD & Cross Entropy & 150/100 & 0.005\\ 
    Resnet-18 & SGD & Cross Entropy & 150/100 & 0.005\\ 
    EfficientNet-B0 & Adam & Cross Entropy & 150/100 & 0.001\\ 
    \bottomrule
    \end{tabular}
    }
    \end{table}

\begin{table}[h!]\centering
    \setlength{\tabcolsep}{.5em}
    \caption{Details for model correction performed for 10 epochs using SGD optimizer with learning rate of $10^{-4}$. We test $\lambda$-values in range $\{1, 5, 10, \dots, 10^4\}$ (\gls{cdep}/\gls{rrr}) or different layers (\gls{clarc}) and pick the best ones on the validation performance.
    Best hyperparameters are given for (ISIC band-aid\,$|$\, synthetic\,$|$\,Bone Age ``L''-marker\,$|$\, synthetic).
    The \gls{rrr} loss is adapted to handle high variance of \gls{dnn} gradients by using cosine similarity (instead of L2-norm) and absolute gradient \wrt the predicted logit.}
    \resizebox{1\textwidth}{!}{
    \begin{tabular}{lccc} \toprule
    {method} & best $\lambda$/layers (VGG-16) &best $\lambda$/layer (ResNet-18) & best $\lambda$/layer (EfficientNet-B0)  \\\midrule
    RRR & 
        $500\,|\,200\,|\,500\,|\,500$ &
         $500\,|\,500\,|\,5000\,|\,1000$ &
        $500\,|\,100\,|\,500\,|\,1000$ 

    \\ 
    CDEP  & 
    \
        $\:10\,|\,1\,|\,100\,|\,5\:\:\:$ &
        $50\,|\,50\,|\,100\,|\,50$ &
        -
    \\ 
    p-ClArC  & 
        \texttt{features} 28\,$|$\,21\,$|$\,26\,$|$\,26 &
        \texttt{layer} $3\,|\,2\,|\,4\,|\,2$ & 
        \texttt{features} $6\,|\,3\,|\,6\,|\,7$  \\  
    a-ClArC  & 
        \texttt{features} 14\,$|$\,21\,$|$\,28\,$|$\,19 &
        \texttt{layer} 4\,$|$\,3\,$|$\,3\,$|$\,2 & 
        \texttt{features} 8\,$|$\,6\,$|$\,2\,$|$\,6 \\ \bottomrule 
    \end{tabular}
    }
    \end{table}


\newpage
\subsection{Additional Results}\label{app:sec:additional_results}

\subsubsection{Artifact Localization}\label{app:sec:additional_results:localization}

\begin{figure}[h] \centering
    \includegraphics[width=0.97\textwidth]{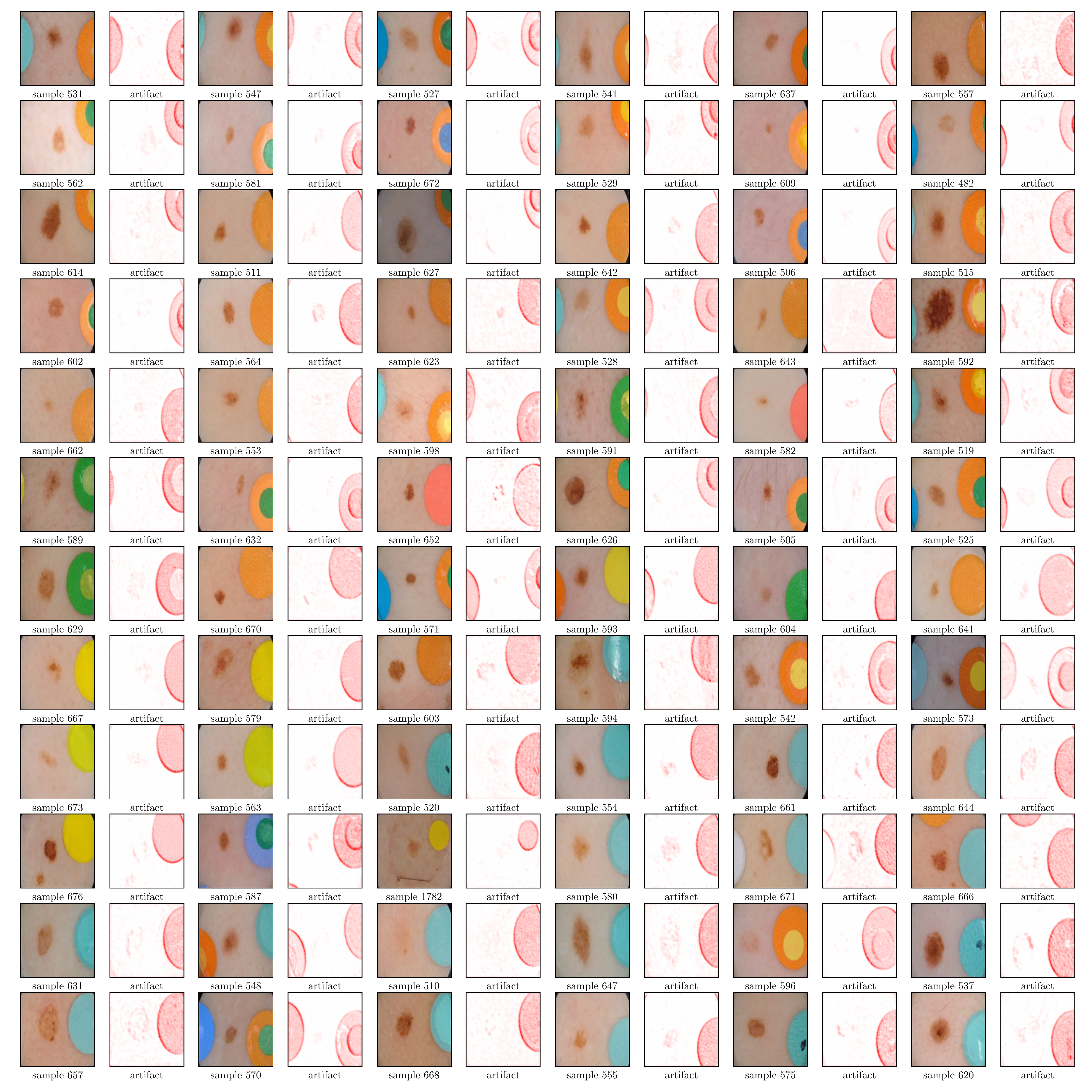}
    \caption{Examples for the \gls{r2r} automatic concept localization scheme using artifact \glspl{cav} for the band-aid artifact of the ISIC dataset. Shown are 36 artifact samples with corresponding \gls{cav} heatmaps. Artifacts have been localized in layer \texttt{features.7} of the VGG-16 model.} 
    \label{app:fig:localization_band_aid}
\end{figure}

\begin{figure}[h] \centering
    \includegraphics[width=0.99\textwidth]{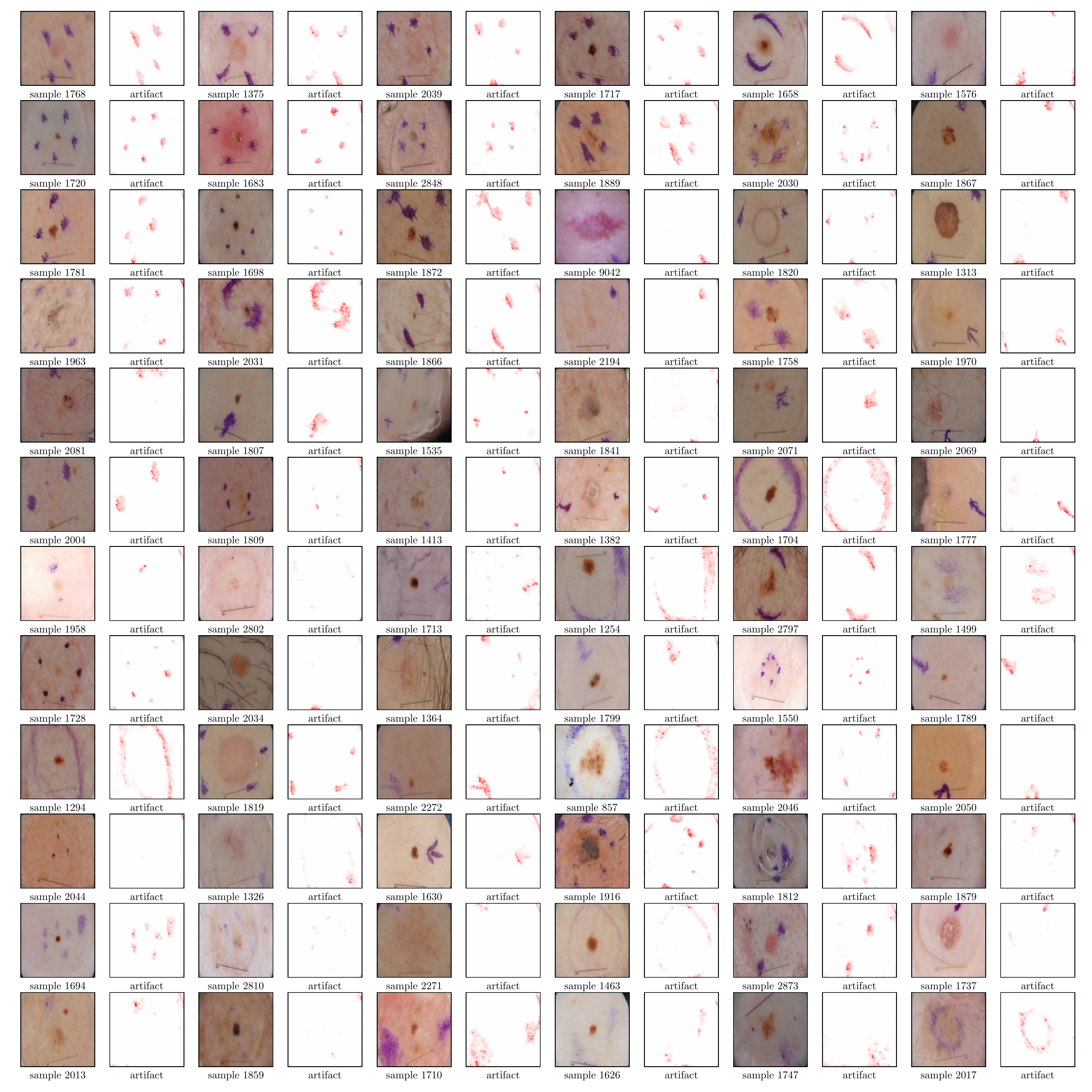}
    \caption{Examples for the \gls{r2r} automatic concept localization scheme using artifact \glspl{cav} for the skin marker artifact of the ISIC dataset. Shown are 36 artifact samples with corresponding \gls{cav} heatmaps. Artifacts have been localized in layer \texttt{features.12} of the VGG-16 model.} 
    \label{app:fig:localization_skin_marker}
\end{figure}

\begin{figure}[h] \centering
    \includegraphics[width=0.99\textwidth]{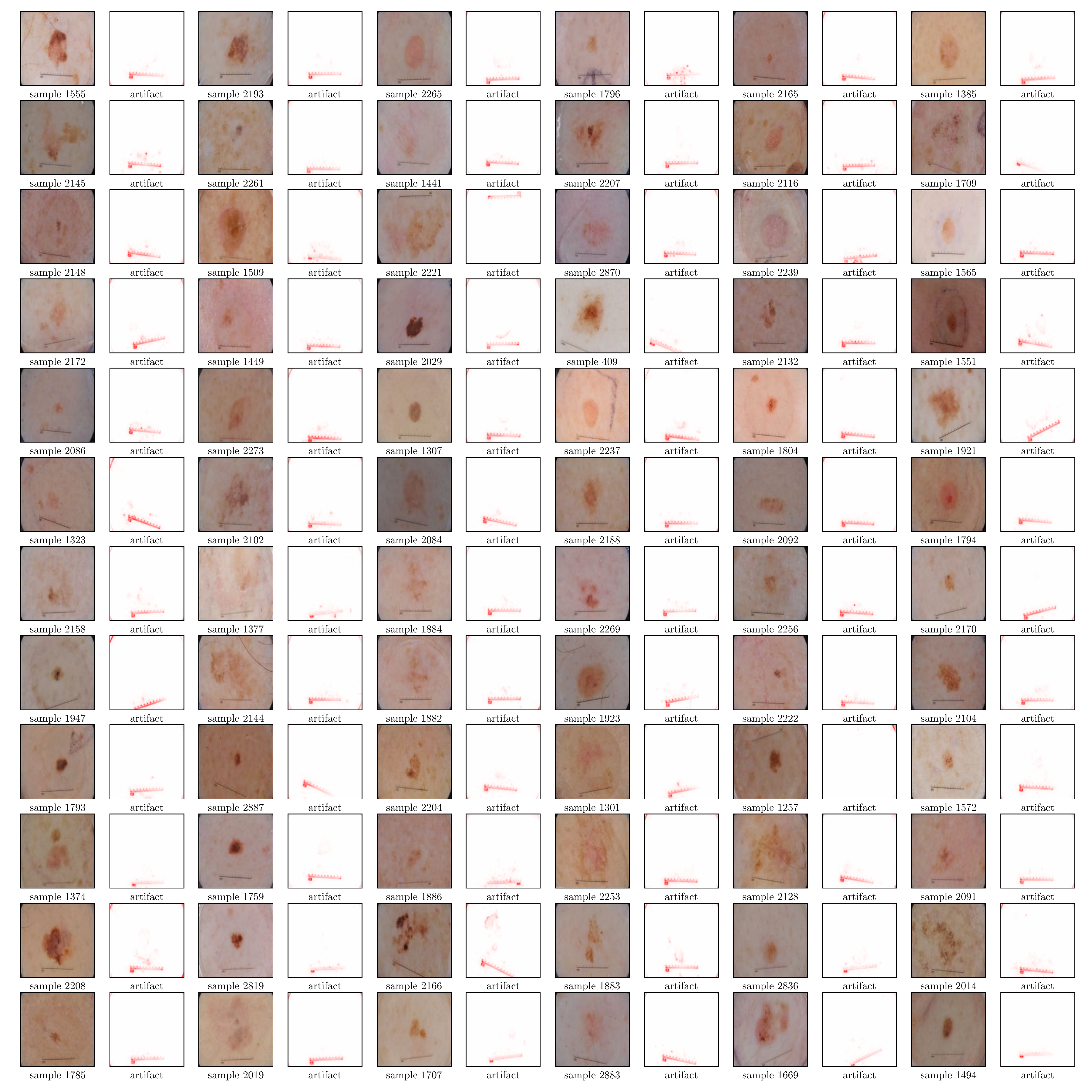}
    \caption{Examples for the \gls{r2r} automatic concept localization scheme using artifact \glspl{cav} for the ruler artifact of the ISIC dataset. Shown are 36 artifact samples with corresponding \gls{cav} heatmaps. Artifacts have been localized in layer \texttt{features.28} of the VGG-16 model.} 
    \label{app:fig:localization_ruler}
\end{figure}

\begin{figure}[h] \centering
    \includegraphics[width=0.99\textwidth]{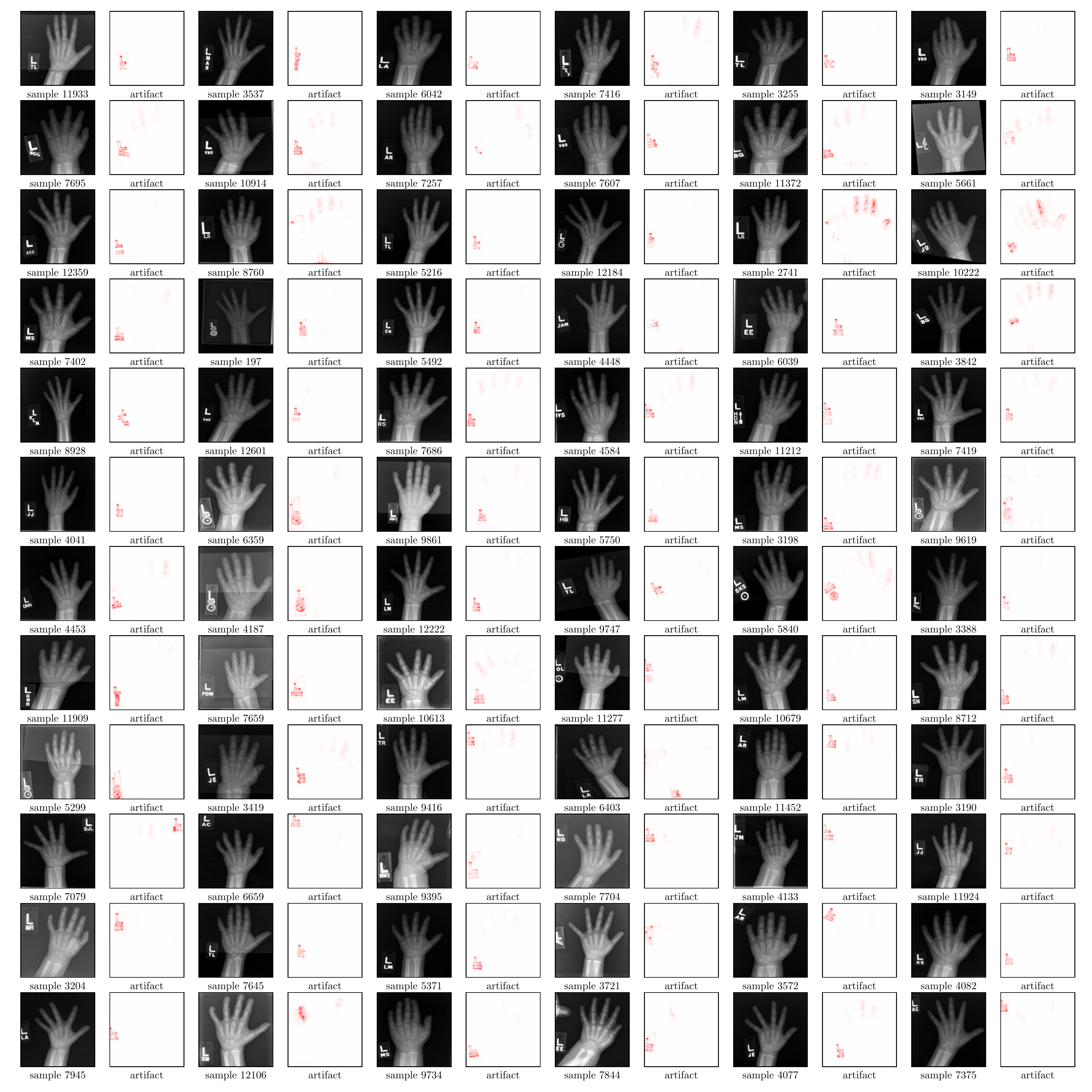}
    \caption{Examples for the \gls{r2r} automatic concept localization scheme using artifact \glspl{cav} for the ``L''-marker artifact of the Bone Age Estimation dataset. Shown are 36 artifact samples with corresponding \gls{cav} heatmaps. Artifacts have been localized in layer \texttt{features.28} of the VGG-16 model.} 
    \label{app:fig:localization_big_l}
\end{figure}

\clearpage
\subsubsection{Model Correction}\label{app:sec:additional_results:model_correction}
        \begin{figure}[!ht] \centering
            \includegraphics[width=0.99\textwidth]{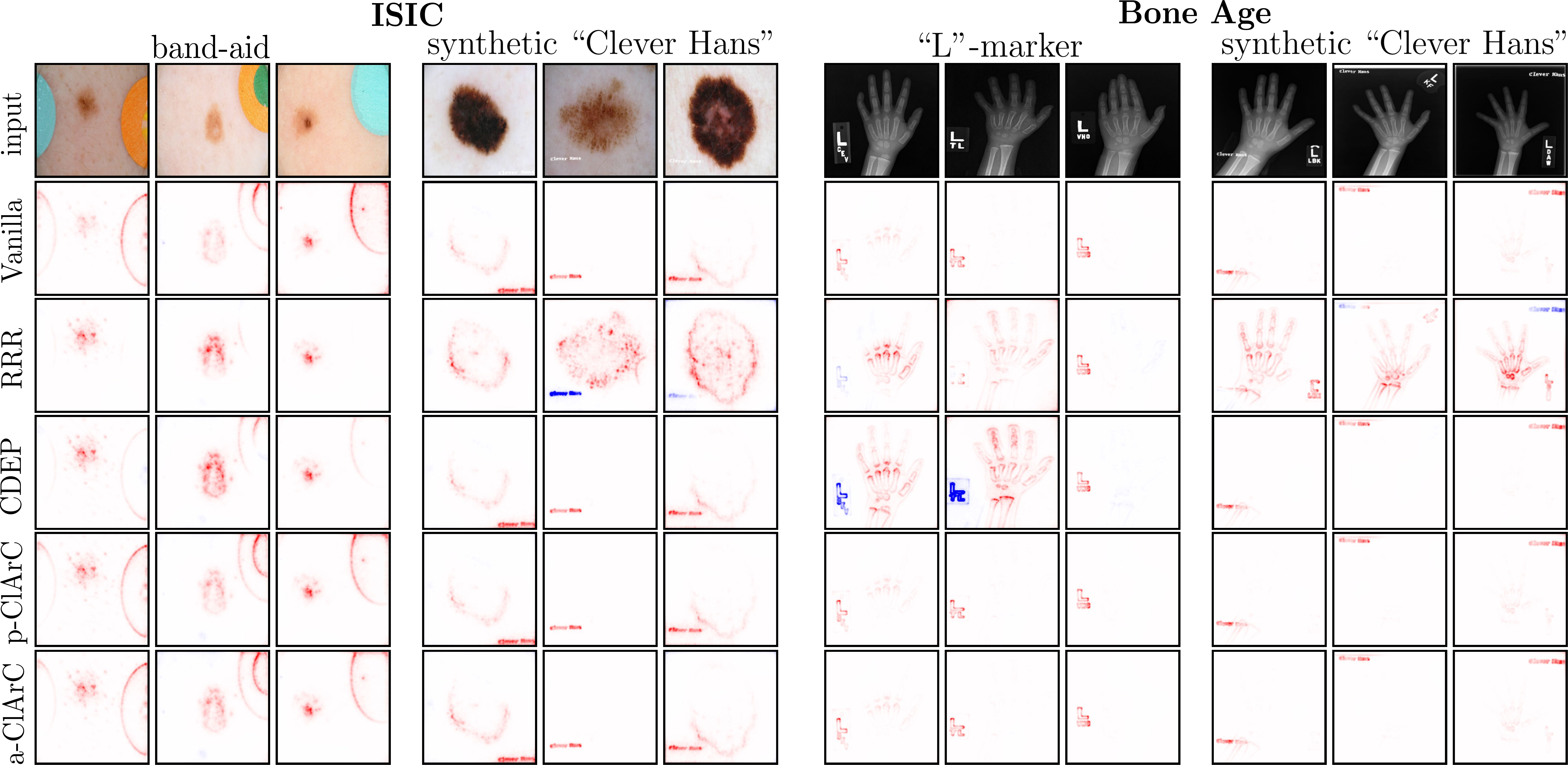}
            \caption{Explanation heatmaps for corrected VGG-16 models using different methods for ISIC2019 (band-aid~$|$~synthetic) and Bone Age (``L''-marker~$|$~synthetic) artifacts.} 
            \label{app:fig:qualititave_correction_results}
        \end{figure}

        \begin{figure}[!ht] \centering
            \includegraphics[width=0.99\textwidth]{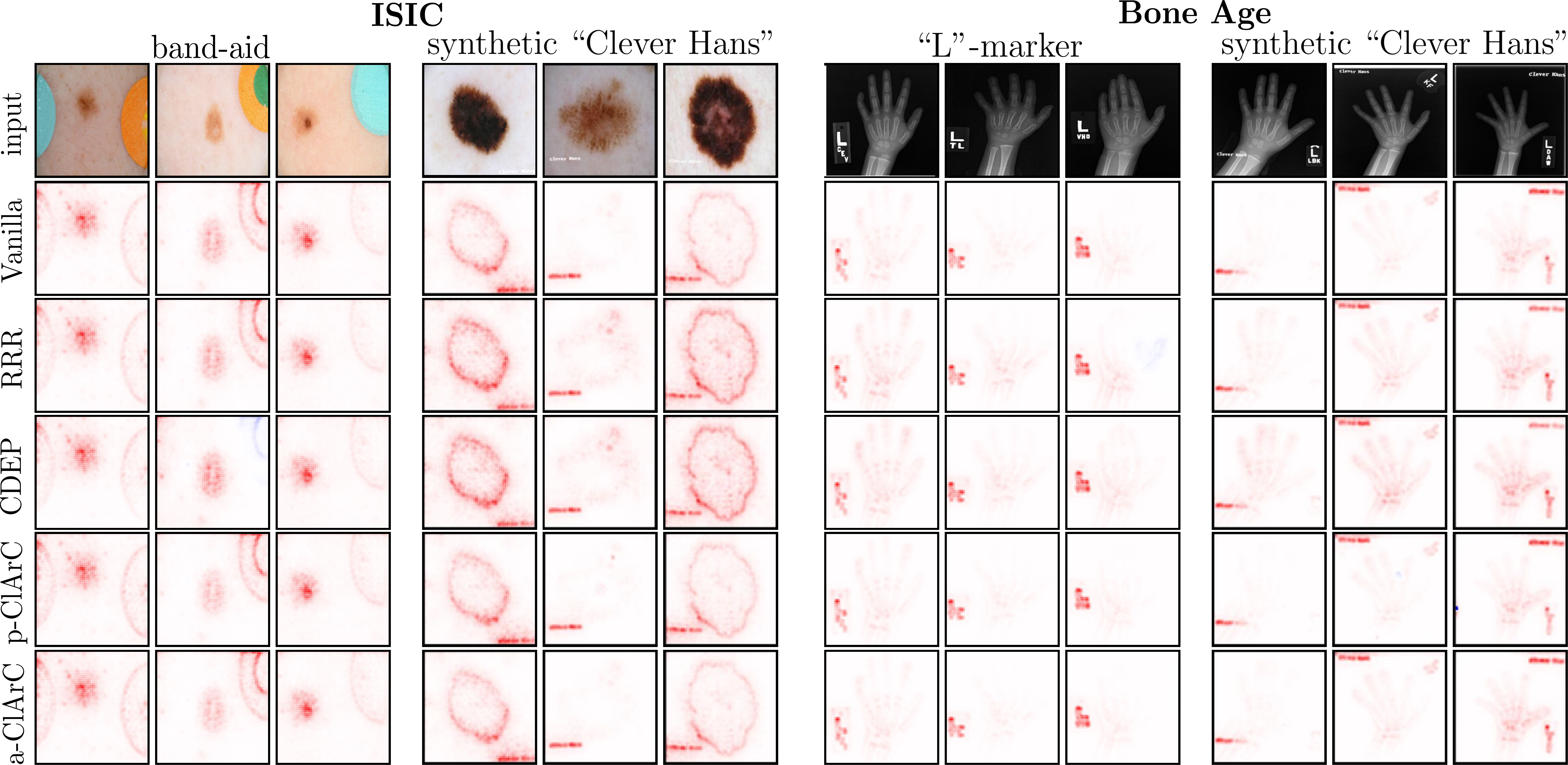}
            \caption{Explanation heatmaps for corrected ResNet-18 models using different methods for ISIC2019 (band-aid~$|$~synthetic) and Bone Age (``L''-marker~$|$~synthetic) artifacts.} 
            \label{app:fig:qualititave_correction_results_resnet}
        \end{figure}
        \begin{figure}[!ht] \centering
            \includegraphics[width=0.99\textwidth]{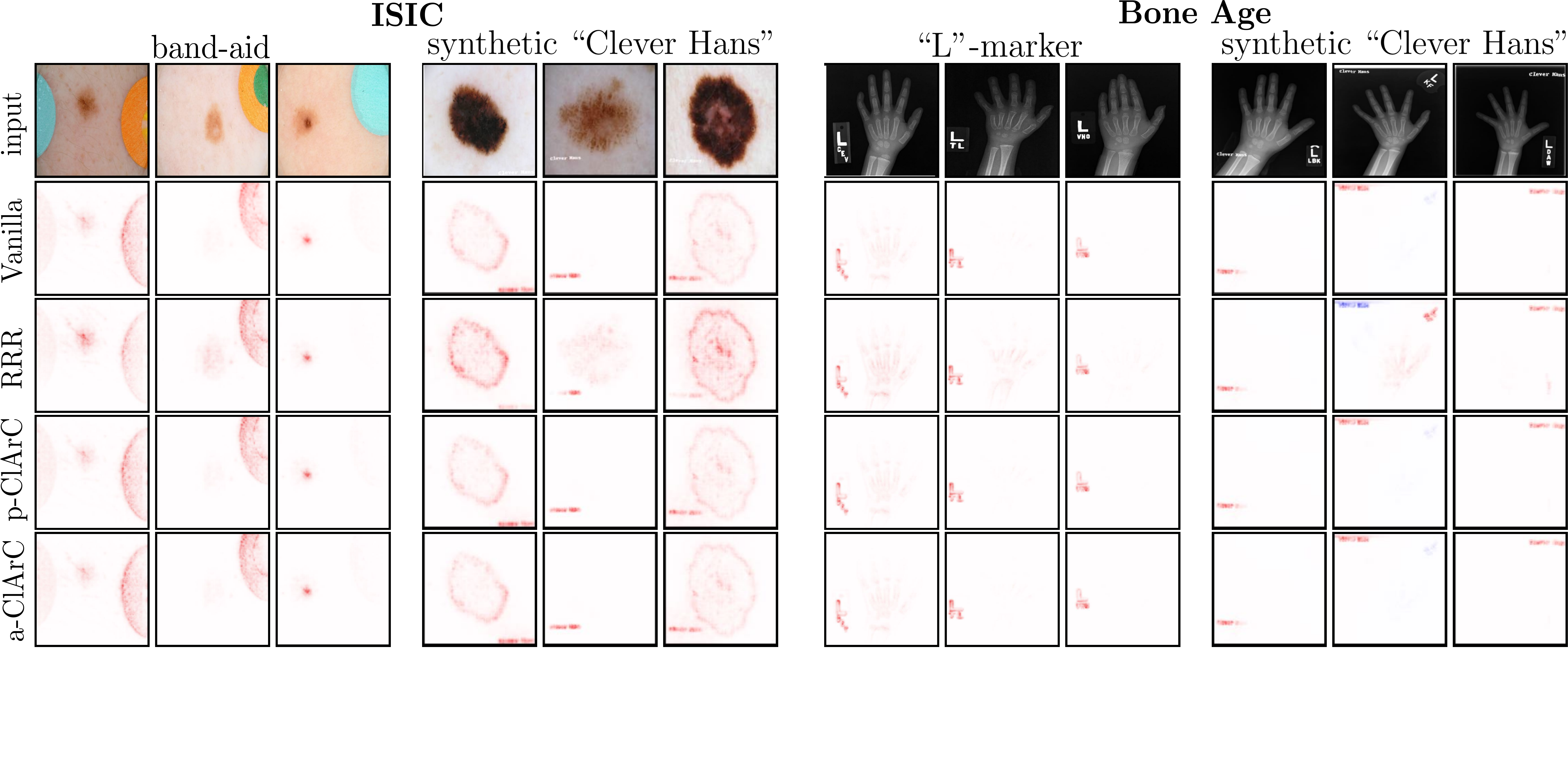}
            \caption{Explanation heatmaps for corrected EfficientNet-B0 models using different methods for ISIC2019 (band-aid~$|$~synthetic) and Bone Age (``L''-marker~$|$~synthetic) artifacts.} 
            \label{app:fig:qualititave_correction_results_efficientnet}
        \end{figure}

    \begin{table}[]\centering
            \caption{Results for the Bone Age Estimation dataset artifacts (``L''-marker~$|$~synthetic). Arrows show whether low ($\downarrow$)/high ($\uparrow$) scores are better with best in bold.}\label{app:tab:model_correction_bone}
    \resizebox{\textwidth}{!}{
    \begin{tabular}{@{}l@{\hspace{1em}}lc@{\hspace{0.5em}}c@{\hspace{0.5em}}c@{\hspace{1em}}c@{\hspace{0.5em}}c@{}}
\toprule
& & $\downarrow$ artifact &  \multicolumn{2}{c}{$\uparrow$ F1 (\%)} &  \multicolumn{2}{c}{$\uparrow$ accuracy (\%)}  \\ 
architecture & method  & relevance (\%) & \textit{poisoned} & \textit{original} & \textit{poisoned} & \textit{original} \\ 
\midrule
\multirow{5}{*}{VGG-16} 
    &       \emph{Vanilla} &                    $\mathbf{}29.6\,|\,\mathbf{}57.5$ &        $\mathbf{}61.4\,|\,\mathbf{}16.2$ &  $\mathbf{}76.6\,|\,\mathbf{}82.3$ &              $\mathbf{}62.2\,|\,\mathbf{}13.6$ &        $\mathbf{}78.6\,|\,\mathbf{}80.3$ \\ \Cline{.1pt}{2-7}\rule{0pt}{2.5ex}
    &           RRR &                      $\mathbf{9.6}\,|\,\mathbf{9.0}$ &        $\mathbf{66.1}\,|\,\mathbf{}44.6\:$ &  $\mathbf{77.4}\,|\,\mathbf{}82.2\:$ &              $\mathbf{65.7}\,|\,\mathbf{}58.5\:$ &        $\mathbf{}78.5\,|\,\mathbf{}80.7$ \\
    &          CDEP &                    $\mathbf{}21.4\,|\,\mathbf{}21.4$ &        $\mathbf{}62.8\,|\,\mathbf{}26.9$ &  $\mathbf{}76.3\,|\,\mathbf{}82.7$ &              $\mathbf{}63.9\,|\,\mathbf{}23.4$ &        $\mathbf{78.7}\,|\,\mathbf{}80.8\:$ \\
    &       p-ClArC &                    $\mathbf{}29.0\,|\,\mathbf{}48.5$ &        $\mathbf{}62.2\,|\,\mathbf{}23.5$ &  $\mathbf{}76.7\,|\,\mathbf{}57.2$ &              $\mathbf{}62.6\,|\,\mathbf{}25.0$ &        $\mathbf{78.7}\,|\,\mathbf{}58.8\:$ \\
    &       a-ClArC &                    $\mathbf{}29.9\,|\,\mathbf{}47.2$ &        $\:\mathbf{}63.2\,|\,\mathbf{55.1}$ &  $\:\mathbf{}76.1\,|\,\mathbf{84.3}$ &              $\:\mathbf{}64.2\,|\,\mathbf{64.7}$ &        $\:\mathbf{}78.1\,|\,\mathbf{81.8}$ \\
    \midrule
\multirow{5}{*}{ResNet-18} 
    &       \emph{Vanilla} &                    $\mathbf{}23.3\,|\,\mathbf{}38.4$ &        $\mathbf{}65.5\,|\,\mathbf{}51.9$ &  $\mathbf{}72.5\,|\,\mathbf{}80.1$ &              $\mathbf{}66.6\,|\,\mathbf{}64.9$ &        $\mathbf{}75.9\,|\,\mathbf{}79.2$ \\ \Cline{.1pt}{2-7}\rule{0pt}{2.5ex}
    &           RRR &                    $\mathbf{16.4}\,|\,\mathbf{}27.7\:$ &        $\mathbf{}67.4\,|\,\mathbf{}59.1$ &  $\mathbf{}71.0\,|\,\mathbf{}79.0$ &              $\:\mathbf{}68.9\,|\,\mathbf{68.0}$ &        $\mathbf{}75.6\,|\,\mathbf{}78.5$ \\
    &          CDEP &                    $\:\mathbf{}19.6\,|\,\mathbf{17.9}$ &        $\mathbf{}65.7\,|\,\mathbf{}56.3$ &  $\mathbf{}71.9\,|\,\mathbf{}77.5$ &              $\mathbf{}66.6\,|\,\mathbf{}64.0$ &        $\mathbf{}74.9\,|\,\mathbf{}76.4$ \\
    &       p-ClArC &                    $\mathbf{}21.4\,|\,\mathbf{}29.2$ &        $\mathbf{}66.5\,|\,\mathbf{}53.7$ &  $\mathbf{72.6}\,|\,\mathbf{79.9}$ &              $\mathbf{}67.7\,|\,\mathbf{}66.5$ &        $\mathbf{76.1}\,|\,\mathbf{79.1}$ \\
    &       a-ClArC &                    $\mathbf{}21.4\,|\,\mathbf{}29.5$ &        $\mathbf{70.0}\,|\,\mathbf{66.2}$ &  $\mathbf{}70.3\,|\,\mathbf{}77.1$ &              $\mathbf{74.4}\,|\,\mathbf{}67.6\:$ &        $\mathbf{76.1}\,|\,\mathbf{}76.1\:$ \\
    \midrule
\multirow{4}{*}{EfficientNet-B0} 
 &       \emph{Vanilla} &                    $\mathbf{}35.1\,|\,\mathbf{}69.1$ &        $\mathbf{}69.5\,|\,\mathbf{}49.4$ &  $\mathbf{}75.0\,|\,\mathbf{}81.6$ &              $\mathbf{}73.6\,|\,\mathbf{}60.3$ &        $\mathbf{}78.1\,|\,\mathbf{}81.4$ \\ \Cline{.1pt}{2-7}\rule{0pt}{2.5ex}
    &           RRR &                    $\mathbf{24.0}\,|\,\mathbf{}51.7\:$ &        $\mathbf{71.2}\,|\,\mathbf{}58.5\:$ &  $\:\mathbf{}74.9\,|\,\mathbf{82.0}$ &              $\mathbf{74.3}\,|\,\mathbf{65.5}$ &        $\mathbf{78.0}\,|\,\mathbf{81.6}$ \\
    &       p-ClArC &                    $\mathbf{}31.1\,|\,\mathbf{}66.2$ &        $\mathbf{}70.4\,|\,\mathbf{}50.4$ &  $\mathbf{}74.9\,|\,\mathbf{}81.7$ &              $\mathbf{}74.0\,|\,\mathbf{}61.1$ &        $\mathbf{}77.9\,|\,\mathbf{}81.3$ \\
    &       a-ClArC &                    $\:\mathbf{}35.3\,|\,\mathbf{24.7}$ &        $\:\mathbf{}70.5\,|\,\mathbf{61.3}$ &  $\mathbf{75.4}\,|\,\mathbf{}78.7\:$ &              $\mathbf{}72.9\,|\,\mathbf{}64.9$ &        $\mathbf{}77.7\,|\,\mathbf{}77.0$ \\
    \bottomrule
\end{tabular}
}
\end{table}

\end{document}